\documentclass[letterpaper]{article} 
\usepackage{aaai24}  

\usepackage{times}  
\usepackage{helvet}  
\usepackage{courier}  
\usepackage[hyphens]{url}  
\usepackage{graphicx} 
\usepackage{natbib}  
\usepackage{caption} 
\usepackage{algorithm}
\usepackage{multirow}
\usepackage{algorithmic}
\usepackage{color}

\usepackage{amsmath,amssymb,amsfonts}
\usepackage{booktabs}

\usepackage{marvosym}
\usepackage{subcaption}
\usepackage{textcomp}
\usepackage{multirow}
\usepackage{algorithm}
\usepackage{algorithmic}

\usepackage{newfloat}
\usepackage{listings}
\DeclareCaptionStyle{ruled}{labelfont=normalfont,labelsep=colon,strut=off} 
\floatstyle{ruled}
\newfloat{listing}{tb}{lst}{}
\floatname{listing}{Listing}
\title{Cell Graph Transformer for Nuclei Classification}
\author{
    Wei Lou\textsuperscript{\rm 1, \rm 2}, 
    Guanbin Li\textsuperscript{\rm 3, \rm 4}, 
    Xiang Wan\textsuperscript{\rm 1}, 
    Haofeng Li\textsuperscript{\rm 1}\thanks{Haofeng Li is the corresponding author.}
}
\affiliations{
    \textsuperscript{\rm 1}Shenzhen Research Institute of Big Data, Shenzhen, China \\
    \textsuperscript{\rm 2}The Chinese University of Hong Kong, Shenzhen, China \\
    \textsuperscript{\rm 3}School of Computer Science and Engineering, Sun Yat-sen University, Guangzhou, China \\
    \textsuperscript{\rm 4}GuangDong Province Key Laboratory of Information Security Technology \\ 
    weilou@link.cuhk.edu.cn,
    liguanbin@cuhk.edu.cn,
    \{wanxiang, lhaof\}@sribd.cn
}

\usepackage{bibentry}

\begin{document}

\maketitle

\begin{abstract}
Nuclei classification is a critical step in computer-aided diagnosis with histopathology images. In the past, various methods have employed graph neural networks (GNN) to analyze cell graphs that model inter-cell relationships by considering nuclei as vertices. However, they are limited by the GNN mechanism that only passes messages among local nodes via fixed edges. To address the issue, we develop a cell graph transformer (CGT) that treats nodes and edges as input tokens to enable learnable adjacency and information exchange among all nodes. Nevertheless, training the transformer with a cell graph presents another challenge. Poorly initialized features can lead to noisy self-attention scores and inferior convergence, particularly when processing the cell graphs with numerous connections. Thus, we further propose a novel topology-aware pretraining method that leverages a graph convolutional network (GCN) to learn a feature extractor. The pre-trained features may suppress unreasonable correlations and hence ease the finetuning of CGT. Experimental results suggest that the proposed cell graph transformer with topology-aware pretraining significantly improves the nuclei classification results, and achieves the state-of-the-art performance. Code and models are available at https://github.com/lhaof/CGT
\end{abstract}
\section{Introduction}
Identifying cell types for histopathology image has emerged as a fundamental task in computational pathology~\cite{krithiga2021breast,amgad2022nucls,huang2023prompt}. 
By effectively classifying nuclei, medical professionals gain crucial insights into the intricate cellular structures, which helps make decisions related to disease diagnosis~\cite{lagree2021review} and 
prognosis~\cite{liu2022pathological}. 
Thus, in this paper, we focus on inferring the types of cell nuclei in a histopathology image.


Deep learning (DL) based methods \cite{graham2019hover,abousamra2021multi,doan2022sonnet} have been widely applied to the nuclei classification task. Most of them employ convolutional neural networks (CNNs) to compute pixel-wise local features and fail to consider the macrostructure of nuclei distribution \cite{anand2020histographs,javed2020cellular}. Another group of methods exploits the cell graph of a histopathology image, and has been studied for decades~\cite{schnorrenberg1996computer,hassan2022nucleus}. 
A \textit{Cell Graph} is a set of vertices and edges, where a vertex is a cell or nucleus and an edge is built between two neighboring cells. Recently, graph convolutional networks (GCNs) have been used to learn embeddings with cell graphs~\cite{zhou2019cgc,zhao2020predicting,anklin2021learning,hassan2022nucleus}. These GCN-based solutions update the embedding of a nucleus by aggregating its adjacent nuclei. However, these GCN methods aggregate features along non-learnable edge connections that are fixed after building a cell graph, which limits the model capacity. 

\begin{figure}[!t]
\includegraphics[width=0.97\linewidth]{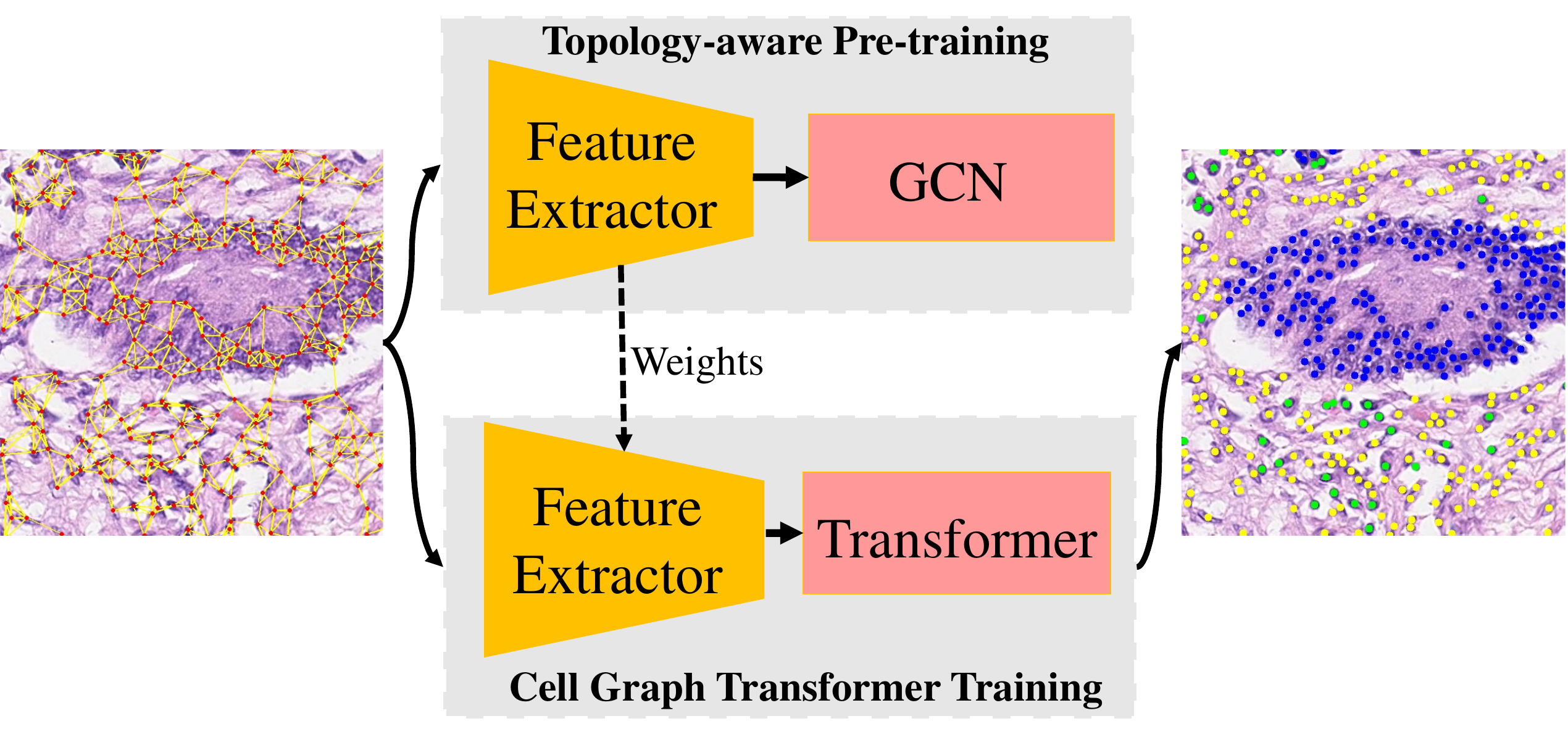}
\centering
\caption{The idea of Topoloyg-aware Pre-training for Cell Graph Transformer. Simple initialization for the cell graph transformer fails to converge due to a large amount of unreasonable connections. The topology-aware pre-training can reduce the initial noise in features and boost the representation ability of the cell graph transformer. 
}
\label{fig1}
\end{figure}

To overcome the issue, we propose a Cell Graph Transformer (CGT) for the nuclei classification task, inspired by \cite{kreuzer2021rethinking,ying2021transformers,kimpure}. The proposed CGT takes both nuclei and edges as input tokens to compute any pairwise correlations among all tokens, which can capture long-range contexts in a more flexible way. CGT is a portable model that can identify cell types, based on any form of binary segmentation or detection results of nuclei. These results could be obtained by existing methods or manual labeling. In the CGT framework, we first compute the centroid coordinates of nuclei from the segmentation/detection result, then determine if two cells are connected by an edge according to their spatial distance, and build up the topology of a cell graph. To obtain visual features for nodes and edges, we develop a U-Net that outputs a pixel-wise feature map with an input image. To embed adjacency into the CGT, we propose a cell-graph tokenization method that integrates the visual and position embeddings of nodes and edges with two kinds of markers, which indicates the type and neighborhood of a token. Afterward, the CGT encoder, which is built of standard transformer layers, performs node-level classification to output the cell types.


Importantly, we observe that simple initialization of the feature extractor fails to train our proposed CGT model. It may be due to that the pairwise attentions can be computed between less correlated tokens, and result in noises especially when the representations are not well initialized. Therefore, we propose a novel topology-aware pretraining strategy that replaces our proposed CGT with a GCN to guide the learning of a feature extractor on the same nuclei classification task. The guiding GCN model only merges the embeddings of adjacent nodes defined by the fixed cell graph, which means less unreasonable correlations and makes convergence easier. The pre-trained feature extractor is supposed to synthesize structure-guided representations that benefit the training of the proposed CGT framework. 

Overall, our contributions have three folds:

\begin{itemize}
     
 \item A nuclei classification framework, Cell Graph Transformer, which benefits from non-local contexts by computing pairwise attentions among all nodes and edges;

\item A topology-aware pretraining strategy that provides topology-guided feature learning for reducing the initial noise of the cell graph transformer;

\item The proposed cell graph transformer significantly surpasses the state-of-the-art methods and our pretraining strategy also brings an improvement to the baseline.
\end{itemize}
\begin{figure*}[!t]
\includegraphics[width=0.85\linewidth]{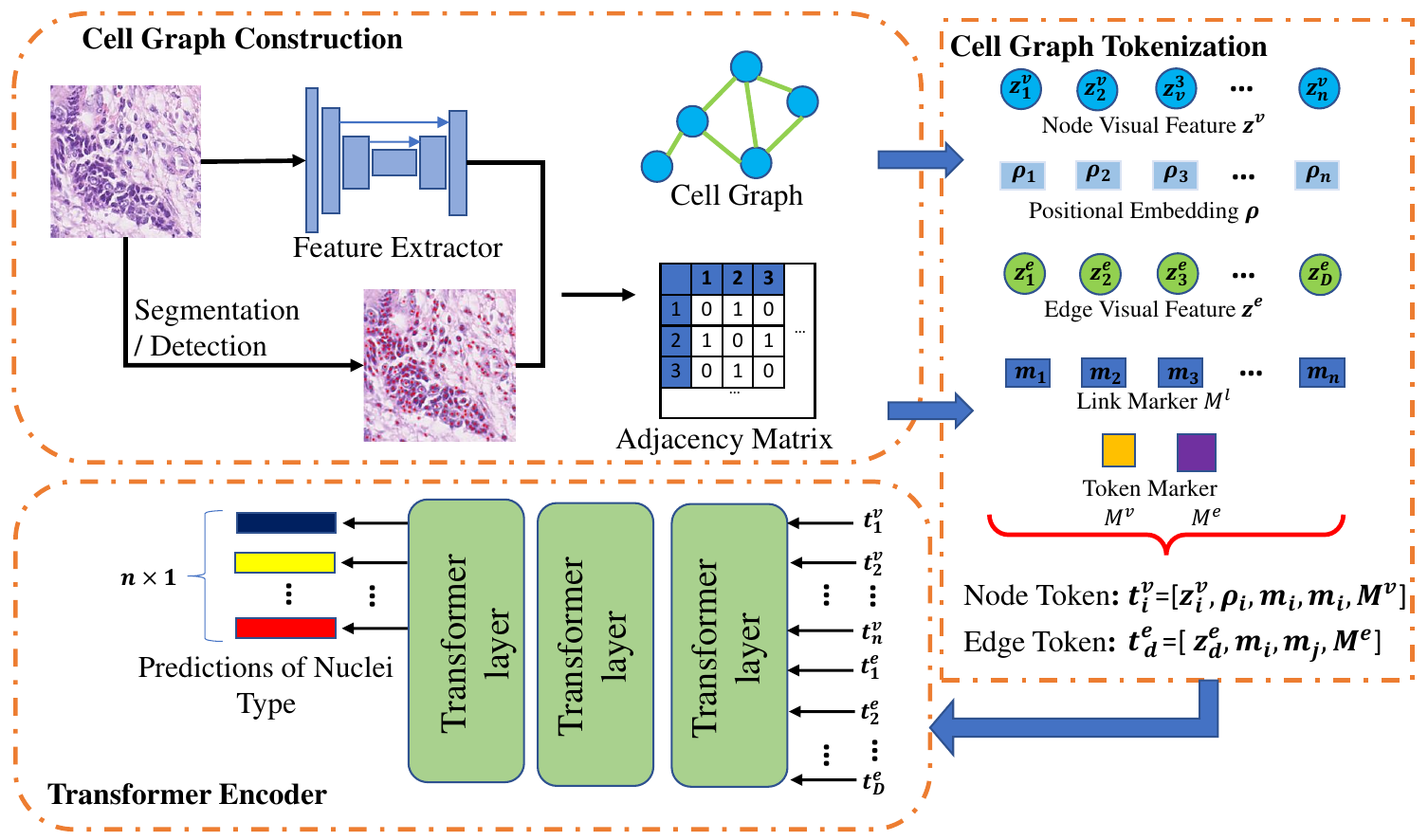}
\centering
\caption{Overview of the Cell Graph Transformer (CGT) framework. The framework includes the construction and tokenization of cell graphs as well as a transformer encoder. Tokens are the input feature vectors of the transformer encoder. Adjacency is embedded into the tokens via link and token markers.  
}
\label{fig2}
\end{figure*}

\section{Related Work}
\noindent\textbf{Nuclei Classification for Histopathology Images.}
Early solutions for nuclei classification involved the extraction of manually defined features 
which are fed into classifiers like SVM or AdaBoost~\cite{liu2011features,sharma2015multi}. However, the handcrafted features limit the representation capabilities of nuclei entities. Recently, the nuclei classification models usually infer cell types based on the CNNs for nucleus segmentation~\cite{zhang2017deeppap,basha2018rccnet,lou2022pixel,lou2023multi,ma2023multi,yu2023diffusion} or nucleus centroid detection~\cite{abousamra2021multi,huang2023affine}.
\citet{graham2019hover} propose a CNN of three branches, predicting nucleus types for the segmented nucleus instances. \citet{doan2022sonnet} incorporated a weight map prediction technique to highlight challenging pixel samples for improved classification. However, these CNN-based approaches are limited by their local pixel-wise receptive field, and fail to capture instance-level contexts among cell nuclei. Therefore, we use a cell graph structure that describes the global relationship among nucleus instances. 

\noindent\textbf{Graph Models in Computational Pathology.}
Graph models have been used in computational pathology for decades. 
\citet{demir2005augmented} builds a graph by considering nuclei as nodes and binary connections as edges. A perceptron is utilized for the detection of inflammation in brain biopsy. 
Recently graph convolution networks (GCNs) have been used as learnable models for the graphs derived from histopathology images~\cite{lou2023structure}. 
Some approaches~\cite{zhou2019cgc,javed2020cellular,pati2022hierarchical} 
classify whole slide images by defining nodes as nucleus instances, superpixels, or tissue patches. In these methods, the node embeddings are hand-crafted or learned features from pre-trained CNN models. NCCD~\cite{hassan2022nucleus} has been proposed for GNN-based nucleus classification recently. However, the GNN-based methods aggregate features along non-learnable edges, which are fixed and limit the model capacity. Thus, we develop a cell graph transformer with learnable node connections and capture the long-range contexts more flexibly. 

\noindent\textbf{Transformers for Graph.}
Transformer models have emerged as crucial components in various domains such as neural language processing~\cite{vaswani2017attention}
and computer vision~\cite{liu2021swin}. Several existing approaches have incorporated transformers to handle graph structures in different manners. First, some methods employ Transformer layers as auxiliary modules within graph neural networks~\cite{wu2021representing,lin2021mesh}. Second, attention matrices are introduced into the message-passing mechanism~\cite{dwivedi2020generalization,zheng2022graph}. 
However, these approaches are constrained by the non-learnable edge connections in the graph structure and may suffer from the issue of excessive smoothing caused by the message-passing mechanism~\cite{li2018deeper,oono2020graph}. Recently, researchers have made progress in graph representation learning by employing pure Transformer architectures with learnable positional encodings~\cite{kreuzer2021rethinking} or by utilizing sparse higher-order Transformers~\cite{kim2021transformers}. 
In this paper, we propose a GCN-guided pretraining strategy that adapts visual features to a graph topology for better training a cell graph transformer on the nuclei classification task.

\section{Methodology}
In this section, we introduce the proposed Cell Graph Transformer framework, the proposed topology-aware pre-training strategy, and the training-inference scheme.

\subsection{Cell Graph Transformer}\label{sec:cgt}

We propose a cell graph transformer (CGT) framework to identify the category of each nucleus in histopathology images. To focus on the classification part, the CGT simply adopts an existing model to provide binary (foreground v.s. background) segmentation/detection results of nuclei. Since CGT performs nuclei classification based on the position coordinates of nuclei, it can adapt to various forms of segmentation/detection results. As Figure~\ref{fig2} shows, our proposed cell graph transformer has three parts: Cell Graph Construction, Cell Graph Tokenization (CGToken), and Transformer Encoder.

\subsubsection{Cell Graph Construction.}
Building a cell graph requires two inputs: 1. Centroid coordinates of nuclei computed by the binary detector/segmentation tool; 2. A feature map $f$ is obtained by the feature extractor in Figure~\ref{fig2}.
Given the centroids of $n$ nuclei in an image, an undirected cell graph $G=(V,E)$ can be constructed by connecting each nucleus centroid to its nearest $k$ nuclei. Therefore, the cell graph contains $n$ nodes $V=\{v_1,...,v_n\}$ and $D(=kn)$ edges $E=\{e_1,...,e_D\}$. The connections of nodes are represented by a binary adjacency matrix $A \in R^{n \times n}$, in which $A_{i,j}$=1 if node $v_i$ and $v_j$ are connected. 

To obtain the visual embeddings of nodes and edges, our approach incorporates a U-Net architecture that leverages an existing convolutional neural network (CNN) \cite{guo2023visual} as the encoder, and a feature pyramid network (FPN) \cite{lin2017feature} as the decoder. The U-Net is initialized via a novel pre-training strategy, which is introduced in the subsequent subsection. Given a histopathology image with dimensions $H \times W$, the U-Net takes the image as input and generates a feature map $f$ of size $\frac{H}{4} \times \frac{W}{4} \times C$ from its second-to-last layer. To describe a nucleus node $v_i$, we first sample the visual embedding $z^v_i \in R^{1 \times C}$ located at the centroid coordinate of the nucleus from $f$. The vector $z^v_i$ can be calculated through bilinear interpolation by using feature vectors at the four nearest integer coordinates on $f$. Besides, we inject spatial positional information by computing a positional embedding vector $\rho_i \in R^{1 \times C}$ using the Sinusoidal Position Encoding method \cite{vaswani2017attention}. The node feature is defined as the concatenation of $z^v_i$ and $\rho_i$. For an edge, the feature vector of its middle point is sampled from the feature map $f$ using bilinear interpolation. The sampled vector is viewed as the edge visual embedding $z^e_d$.

\begin{figure*}[!t]
\includegraphics[width=0.9\linewidth]{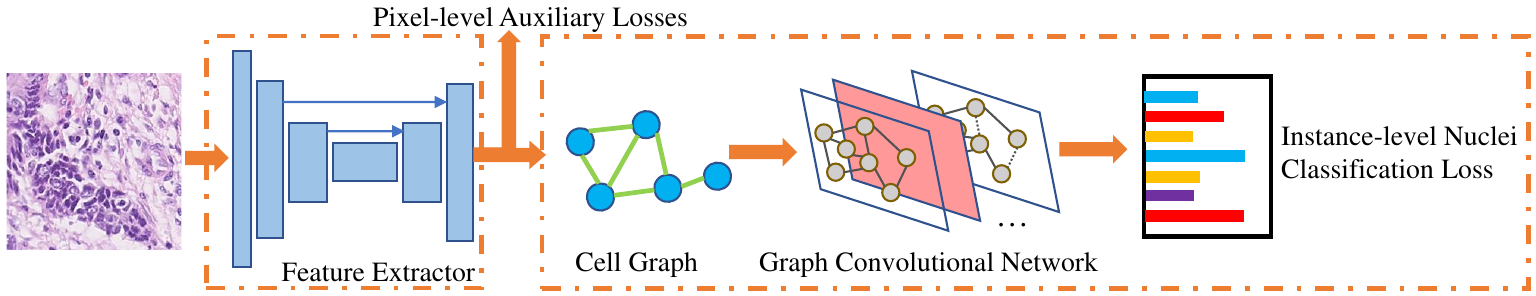}
\centering
\caption{The proposed Topology-Aware Pretraining Strategy. It trains the feature extractor with an instance-level nuclei classification loss and two pixel-level auxiliary losses. The model weights of the trained feature extractor are then used to initialize the feature extractor in training the proposed cell graph transformer.}
\label{fig3}
\end{figure*}

\subsubsection{Cell Graph Tokenization.}
Tokenization is to convert raw data into meaningful numerical representations called \textit{Tokens} that can be well encoded by transformers. We introduce the Cell Graph Tokenization approach (CGToken), which aims to translate the constructed cell graph into a set of tokens that standard transformer models can effectively process. 
%
It is straightforward to regard the node and edge embeddings as $(n+D)$ independent inputs, but it overlooks the structural information contained within the graph. To exploit the topology structure, we utilize \textit{Link Markers} and \textit{Token Markers}. A cell graph transformer framework has two token markers denoted as $M^v,M^e$, which are two $1 \times C$ vectors and the learnable parameters of the framework. One is for node tokens, while the other is for edges. The two token markers are tuned in the training and fixed after training. 

The link markers $M^l = \{m_1,m_2,...,m_n \} \in R^{n \times c}$ are orthonormal vectors that imply the adjacency of each token. If $v_i$ and $v_j$ are linked by an edge, then $[m_i,m_j][m_i,m_i]^T=1$ and $[m_i,m_j][m_j,m_j]^T=1$, otherwise the dot production is 0. This mechanism makes the transformer assign more attention to the nodes connected in the cell graph. The link markers are calculated by the Laplacian eigendecomposition~\cite{dwivedi2020benchmarking} of the adjacency matrix $A$:
\begin{align}\label{eq:lap}
{L}=I-\Theta^{-\frac{1}{2}} A \Theta^{-\frac{1}{2}}=(M^l)^T\alpha M^l, 
\end{align}
where $\Theta$ is the degree matrix of the graph, $I$ is an identity matrix. $\alpha$ and $M^l$ are the eigenvalues and eigenvectors, respectively.  
Then, we further define each node/edge token ($t^v_i$ / $t^{e}_d$) using node/edge visual feature, link and token makers:
\begin{align}\label{eq:tokens}
\begin{array}{l}t^v_i=\left[\sigma_1([z^v_i,\rho_i]),  \sigma_3([m_i, m_i]), M^v\right], i \in\{1,\cdots,  n\},\\
	t^{e}_d=\left[\sigma_2(z^e_d), \sigma_3([m_i, m_j]), M^e\right], d \in\{1,\cdots, D\}, 
\end{array}
\end{align}
where $z^v_i$ and $\rho_i$ are the visual and positional embeddings of the $i^{th}$ node. $z^e_d$ is the visual feature of the $d^{th}$ edge that connects the $i^{th}$ and $j^{th}$ nodes. $[\cdot]$ denotes the concatenation operator. $\sigma_1,\sigma_2,\sigma_3$ are linear projection layers that convert the dimension of their inputs to the same dimension $C$. After computing Eq.~(\ref{eq:tokens}), we obtain $n$ node tokens and $D$ edge tokens. Each token is a vector of size $1\times 3C$.

\subsection{Inference and Training Scheme}
Given the $(n+D)$ tokens, a linear projection layer converts each of these tokens into a $C$-dimensional vector separately. The resulting $(n+D)$ vectors are then fed into the transformer encoder. We employ the standard transformer architecture \cite{vaswani2017attention} for each transformer layer in the encoder. Each layer is composed of a stack of multi-head self-attention layers and a feed-forward network.
To classify the categories of $n$ nodes, we only select the first $n$ features $O \in R^{n \times C}$ from the output of the CGT encoder and send these features into the classification layer. The classification layer is built of a fully-connected (FC) layer and a Softmax function: $P = Softmax(\sigma(O))$. 

Before training the CGT encoder, we pretrain the feature extractor using our proposed topology-aware strategy described in the next subsection. In the training stage, the feature extractor is also fine-tuned with the transformer encoder synchronously. The classification loss of each nucleus node has a cross-entropy term and a focal loss term, and is defined as:
\begin{align}\label{eq:cls_loss}
\mathcal{L}(P,y) = - \sum_{b=1}^{B} y_{b} \log P_{b} - \sum_{b=1}^{B} \tau_{b} (1-P_{b})^\gamma y_{b} \log P_{b},
\end{align}
%
%
where $\gamma$ is a hyper-parameter set to 2, $B$ is the number of categories, $y$ is the true label, $P$ is the prediction and $\tau_{b}$ is the loss weight computed as the reciprocal of the proportion of the $b^{th}$ class in the training set.

\begin{table*}[!thb]
\centering

\setlength{\tabcolsep}{1.8pt}
\begin{tabular}{c|ccccccccc|cccccccc}
\hline
\multirow{2}{*}{Method} & \multicolumn{8}{c}{PanNuke} & &\multicolumn{8}{c}{NuCLS}                                             \\ \cline{2-18}
   & $AJI$   & $PQ$    & $F_d$  & $F^{i}$&$F^{c}$&$F^{d}$&$F^{ep}$&$F^{ne}$  & $\boldsymbol{F_{avg}}$ & $AJI$   & $PQ$    & $F_d$  & {$F^t$}  & {$F^{st}$}  & {$F^s$}  & {$F^o$} & $\boldsymbol{F_{avg}}$ \\ \hline

MCSPat   & -     & -     & 0.786 & 0.484 & 0.473 & 0.220 & 0.612 & 0.629 & 0.514  & -     & -     &  0.658   & 0.488 & 0.267 & 0.581 & 0.035 & 0.343 \\ 
Mask2former  & 0.616 & 0.666 & 0.792 & 0.400 & 0.426 & 0.289 & 0.668 & 0.617 & 0.480  & 0.229 & 0.331 &  0.432   & 0.367 & 0.098 & 0.521 & 0.000 & 0.247 \\ 
SONNET   & 0.686 & 0.649 & 0.813 & 0.522 & 0.474 & \textbf{0.367} & 0.639 & 0.604 & 0.521  & 0.332 & 0.403 & 0.458    & 0.461 & 0.181 & 0.547 & 0.000 & 0.330 \\ 
NCCD & - & - & 0.800 & \textbf{0.571} & 0.525 & 0.354 & 0.660 & 0.588 & 0.539 & - & - & -    & - & - & - &-  & - \\ \hline
Hover.  & 0.663 & 0.631 & 0.793 & 0.510 & 0.478 & 0.265 & 0.627 & 0.636 & 0.503 & 0.467 & 0.429 & 0.662  & 0.469 & 0.272 & 0.586 & 0.023 & 0.337 \\
Ours+Hover.       & 0.663 & 0.631 & 0.793 & 0.527 & \textbf{0.531} & 0.358 & \textbf{0.705} & \textbf{0.673} & \textbf{0.558}  & 0.467 & 0.429 & 0.662    & \textbf{0.501} & \textbf{0.300} & \textbf{0.593} & \textbf{0.095} & \textbf{0.377} \\
Ours+GT  & - & - & - & 0.618 & 0.661 & 0.452 & 0.741 & 0.806 & 0.656  & - &- &- &0.785 & 0.466 & 0.733  & 0.123 & 0.527 \\
\hline
\end{tabular}
\caption{Comparison with the state-of-the-art methods on PanNuke and NuCLS datasets. The best classification results are in bold. `Hover.' and `MCSPat' denotes Hover-net and MCSPatnet. `Ours+Hover.' and `Ours+GT' denote our CGT framework using the segmentation masks from a trained Hover-net model or ground truth.}
\label{table_sota1}
\end{table*}

\begin{table*}[!thb]
\centering

\setlength{\tabcolsep}{1.6pt}
\begin{tabular}{c|cccccccccc|c|ccccc}
\hline
\multirow{2}{*}{Method} & \multicolumn{9}{c}{Lizard} & & \multirow{2}{*}{Method} & \multicolumn{5}{c}{BRCA-M2C}                                            \\ \cline{2-11} \cline{13-17}
   & $AJI$   & $PQ$    & $F_d$  & $F^{ne}$&$F^{ep}$&$F^{l}$&$F^{p}$&$F^{e}$  & $F^{c}$ & $\boldsymbol{F_{avg}}$  & & $F_d$  & $F^{i}$&$F^{ep}$&$F^{s}$& $\boldsymbol{F_{avg}}$                                  \\ \hline

MCSPat.   & -     & -     & 0.705 & 0.110 & 0.604 & 0.457 & 0.228 & 0.210 & 0.478  & 0.347 & DDOD & 0.585 & 0.379 & 0.540 & 0.156 & 0.359 \\
Mask2former  & 0.385 & 0.297 &  0.603 & 0.036  & 0.469& 0.367 & 0.148 & 0.268 & 0.275  & 0.313 & YOLOX & 0.638 & 0.439 & 0.502 &  0.170 & 0.370  \\
SONNET   & 0.434 & 0.447 & 0.597 & 0.197 & 0.610 & 0.322 & 0.328 & 0.402 & 0.421  & 0.380  & ConvN.Uper. & 0.785 & 0.423 & 0.636 & 0.353 & 0.471 \\
NCCD & - & - & 0.633 & \textbf{0.378} & 0.423 & 0.404 & \textbf{0.471} & \textbf{0.461} & 0.534 & 0.445  & DINO & 0.633 & 0.403 & 0.631 & 0.213 & 0.416 \\ \hline
Hover.  & 0.463 & 0.460 & 0.732 & 0.221 & 0.693 & 0.447 & 0.369 & 0.387 & 0.493 & 0.435 & MCSPat. & 0.831 & 0.422 & 0.683 & 0.417 & 0.507 \\ 
Ours+Hover.       & 0.463 & 0.460& 0.732 & 0.302 & \textbf{0.724} & \textbf{0.438} & 0.434 & 0.416 & \textbf{0.548} & \textbf{0.477} & Ours+MCSPat & 0.831 & \textbf{0.447} & \textbf{0.732} & \textbf{0.428} & \textbf{0.536}   \\
Ours+GT      & - & - &  - & 0.508 & 0.868 & 0.543 & 0.537 & 0.585 & 0.678 & 0.620 & Ours+GT. & - &  0.598 & 0.869 & 0.602 & 0.690   \\
\hline
\end{tabular}
\caption{Comparison with the state-of-the-art methods on Lizard and BRCA-M2C datasets.  `ConvNUper.' denotes the ConvNext-UperNet. Since the BRCA-M2C is a nuclei detection and classification benchmark, several nuclei detection and classification methods are utilized for comparison. The best classification results are in bold.}
\label{table_sota2}
\end{table*}

\subsection{Topology-Aware Pretraining Strategy}\label{sec:pretrain}
In the proposed CGT, we find that the initial visual features of nodes/edges play a crucial role in model training. At the early training stage, if visual features are not well initialized, computing correlations of all pairs of node/edge tokens may produce unreliable attention, bringing noise into the CGT training. Graph convolutional network (GCNs) and Transformers have their own strengths in modeling graph data. GCNs excel at capturing local structural information and propagating it across the graph. We consider that the message passing in GCNs is locally guided by fixed edges and is more robust at the start of training. Thus, we propose a pretraining strategy that employs a GCN to help learn the feature extractor in advance.  After that, the cell graph transformer is initialized with the GCN-guided representations, which can help the transformer converge faster and improve the final classification performance.

As shown in Fig.~\ref{fig3}, the proposed pretraining strategy trains a feature extractor with an instance-level nuclei classification loss and two auxiliary pixel-level losses (including the Dice and Cross-entropy losses). The auxiliary losses are for semantic segmentation, and the segmentation result is predicted from the last layer of the feature extractor. 
Note that these two segmentation losses aim at learning the feature extractor instead of producing nucleus masks. The segmentation mask of nuclei is obtained via an existing segmentation tool as shown in Fig.~\ref{fig2}. To compute the instance-level loss in Fig.~\ref{fig3}, the second-last layer of the feature extractor yields a feature map to build the node/edge features of a GCN~\cite{li2021deepgcns}. The cell graph, the edges and their features are defined as the same as those in our proposed CGT. For nuclei segmentation and classification datasets~\cite{gamper2020pannuke,graham2021lizard}, we define the node features to exploit the segmentation masks, following Wei et al.~\citeyearpar{Wei2023Structure}. After that, both the node and edge features are input to the GCN for node update. The enhanced node embeddings are fed into a linear classifier to predict nucleus types. The instance-level nuclei classification loss is the same as Eq.~(\ref{eq:cls_loss}), by viewing $P$ as the GCN prediction. In the pretraining, the feature extractor and the GCN are end-to-end tuned.

\section{Experiments}
\noindent{\textbf{Datasets.}}
We utilize four nuclei classification datasets: PanNuke \cite{gamper2020pannuke}, Lizard \cite{graham2021lizard}, NuCLS \cite{amgad2022nucls}, and BRCA-M2C \cite{abousamra2021multi}. PanNuke, Lizard and NuCLS have the segmentation masks of nuclei, while BRCA-M2C only provides centroid annotations for nuclei detection and classification. 
The PanNuke dataset comprises 7901 images with a size of $256\times 256$ from 19 organs, which includes the cell types of inflammatory, connective, dead, epithelial, and neoplastic. The Lizard benchmark consists of 291 large images with an average size of $1016 \times 917$, which is composed of six existing datasets: ConSeP \cite{graham2019hover}, CRAG, GLAS \cite{sirinukunwattana2017gland}, DigestPath, TCGA \cite{grossman2016toward}, and PanNuke. Lizard contains the nucleus types of epithelial, lymphocyte, plasma, neutrophil, eosinophil, and connective. The NuCLS dataset has 1744 image patches that are grouped into four superclasses: tumor, stromal, sTILs, and other. The BRCA-M2C dataset includes 120 image patches collected from TCGA, has the cell types of inflammatory, epithelial, and stromal. The data split and more details are in the supplementary material.

\begin{table*}[!htb]
\centering

\setlength{\tabcolsep}{3.0pt}
\begin{tabular}{c|ccccccccc}
\bottomrule
Models & Initializing Feature Extractor & Classifier    & $F^{i}$&$F^{c}$&$F^{d}$&$F^{ep}$&$F^{ne}$  & \textbf{$F_{avg}$}  \\ \hline
M1 & UNet ImageNet pretrained & Linear  & 0.510 & 0.463 & 0.065 & 0.668 & 0.000 & 0.341 \\ 
M2 & UNet ImageNet pretrained & Transformer  & 0.456 & 0.381 & 0.166 & 0.601 & 0.614 & 0.444 \\ 
M3 & UNet ImageNet pretrained & GCN      & 0.513 & 0.506 & 0.263 & 0.674 & 0.633 & 0.518 \\ \hline 

M4 & UNet ImageNet pretrained & CGToken+Transformer              & 0.435 & 0.104 & 0.057 & 0.526 & 0.602 & 0.344 \\ 
M5 & UNet Linear pretrained & CGToken+Transformer                & 0.518 & 0.472 & 0.138 & 0.674 & 0.237 & 0.420 \\
M6 &  UNet Transformer pretrained  &   CGToken+Transformer       & 0.484 & 0.438 & 0.237 & 0.634 & 0.626 & 0.484\\ 
M7 & UNet GCN pretrained & Transformer   & 0.525  & 0.484 & 0.228 & 0.672 & 0.655 & 0.512 \\ 
M8 (Ours) & UNet GCN pretrained & CGToken+Transformer   & \textbf{0.527} & \textbf{0.531} & \textbf{0.358} & \textbf{0.705} & \textbf{0.673} & \textbf{0.558} \\ \bottomrule
\end{tabular}
\caption{Ablation study on PanNuke dataset. `CGToken+Transformer' is the proposed classifier in our CGT framework. All the results are based on the official data split of the PanNuke dataset. 
 The best results are in bold.}
 
\label{table_abl}
\end{table*}

\noindent{\textbf{Implementation details.}}
The implementation is based on PyTorch \cite{paszke2017automatic} and PyTorch Geometric library \cite{fey2019fast}. 
For the proposed CGT, the encoder and decoder of the feature extractor have four layers and three layers, respectively. The CGT encoder contains four transformer layers. For the pretraining strategy, the GCN  is built of two GENConv~\cite{li2020deepergcn} layers. Our results are reported as the average result of training with three different random seeds. The dimensions of type markers and link markers are 64 and 16. The number of edges of each node is 4. The pretraining strategy and the training of CGT are run for 150 and 50 epochs, respectively, with the Adam optimizer in an NVIDIA A-100 GPU. The initial learning rates for pretraining and training are $10^{-4}$ and $10^{-5}$, respectively. The overall training time is 2 days for each dataset.

\noindent{\textbf{Metrics.}}
We utilize F-score \cite{graham2019hover} for evaluating classification performance. $F^{i}$, $F^{c}$, $F^{d}$, $F^{ep}$, $F^{ne}$, $F^{t}$, $F^{st}$, $F^{s}$, $F^{o}$, $F^{n}$, $F^{l}$, $F^{p}$, $F^{e}$ denote the class-wise F-score for inflammatory, connective, dead, epithelial, neoplastic, tumor, stromal, sTIL, other, neutrophil, lymphocyte, plasma, eosinophil, respectively. $F_{avg}$ denotes the average F-score for all classes in the same dataset. For evaluating segmentation and detection, we adopt Aggregated Jaccard Index ($AJI$) \cite{Mahmood2019Deep}, Panoptic Quality ($PQ$) \cite{kirillov2019panoptic}, and Detection Quality ($F_d$)~\cite{graham2019hover}. 

\begin{table}[!tb]
\centering
\setlength{\tabcolsep}{2.0pt}
\begin{tabular}{c|c|c|c}
\bottomrule
Method & \#Para. (M) & Infer Time (s) & Model Size (Mb) \\ \hline 
Hover-net & 33.60 & 1799 &  144        \\ 
Ours &  37.43 & 447 &  465       \\ \bottomrule
\end{tabular}
\caption{Computational efficiency on whole slide images. Inference time is measured as the average time of inferring ten whole slide images.}
\label{tab_compute}
\end{table}

\subsection{Comparison with the State-of-the-art Methods}
\begin{figure*}[!t]
\includegraphics[width=0.99\linewidth]{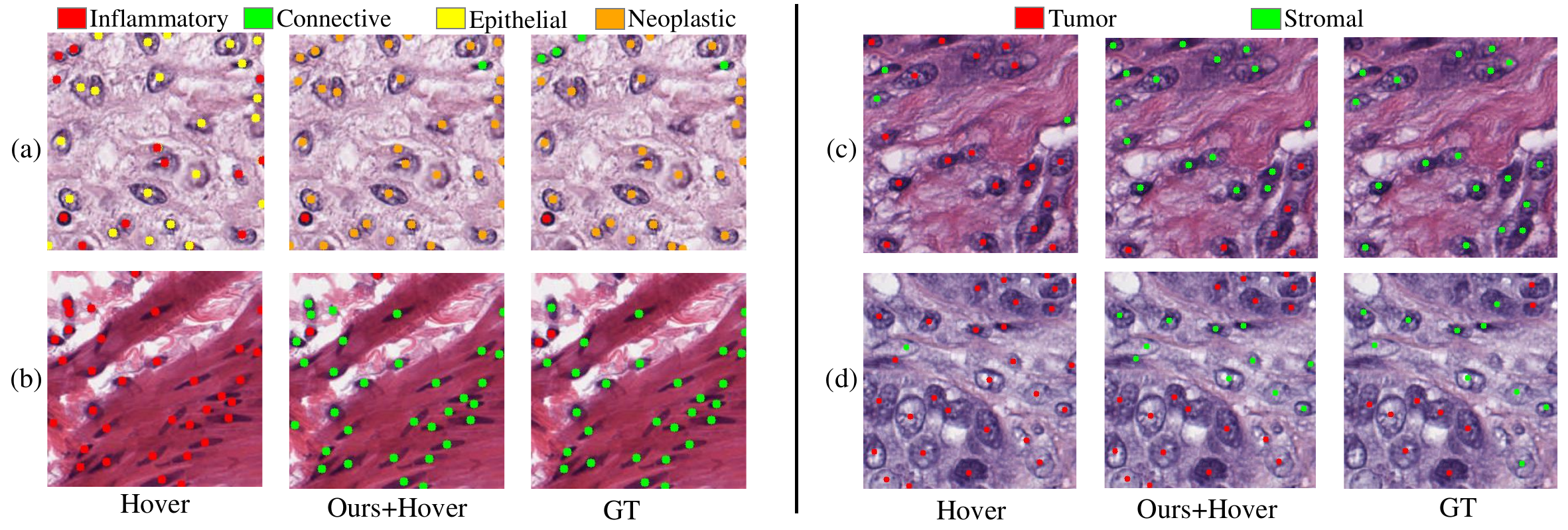}
\centering
\caption{Visual comparison of the proposed CGT with Hover-net on PanNuKe (left) and NuCLS (right) datasets. GT denotes the ground truth.}
\label{fig_sotavisual}
\end{figure*}

For PanNuke, Lizard and NuCLS datasets, the proposed CGT is compared with existing methods: Hover-net \cite{graham2019hover}, MCSPatnet \cite{abousamra2021multi}, SONNET \cite{doan2022sonnet}, Mask2former \cite{cheng2022masked}, NCCD \cite{hassan2022nucleus}. Among them, Hover-net, SONNET and Mask2former are nuclei segmentation and classification methods, MCSPatnet is a nuclei detection and classification method and NCCD is a pure nuclei classification method. For BRCA-M2C dataset, we compare the proposed method with nuclei detection and classification methods: DDOD \cite{chen2021disentangle}, YOLOX \cite{ge2021yolox}, ConvNext-UperNet \cite{liu2022convnet}, MCSPatnet \cite{abousamra2021multi} and DINO \cite{zhang2022dino}. 
In Table \ref{table_sota1} \& \ref{table_sota2}, `Ours+Hover-net' or `Ours+MCSpat' indicates that our CGT utilizes Hover-net or MCSPatNet to generate nuclei segmentation or detection results, without using the predictions of cell types. 
The numerical results of SONNET and NCCD are collected from their papers. As Table~\ref{table_sota1} shows, our proposed method  `Ours+Hover-net' outperforms the second best models by 1.9\%, 3.4\% and 3.2\% in $F_{avg}$ on PanNuke, NuCLS and Lizard, respectively. `Ours+MCSPat.' surpasses the second-best model by 2.9\% in $F_{avg}$ on the BRCA-M2C dataset. Figure~\ref{fig_sotavisual} presents a visual comparison between our proposed CGT and Hover-net on two datasets. Both methods employ the same segmentation masks, but our method shows more accurate classification results of nuclei. More visual results can be found in the appendix.

\noindent\textbf{Effectiveness and Generalization of CGT.} Note that the segmentation tools used in our CGT can also produce classification results. Thus, the CGT is compared with them to show its strength. Comparing `Ours+Hover' with Hover-net suggests that the CGT brings a significant improvement of 1.9\%-3.4\% in the average F-score on three benchmarks. Importantly, on the BRCA-M2C dataset, the proposed CGT also boosts MCSPatNet by 2.9\% in $F_{avg}$ to achieve the state-of-the-art performance. The improvements with two different segmentation/detection tools verify the generalization of our CGT framework. 

In Table \ref{table_sota1} \& \ref{table_sota2}, `Ours+GT' means that our CGT accesses the ground truth of binary segmentation in the testing. It shows that with more accurate nuclei centroids, our CGT can produce better classification results. We claim that the proposed cell graph transformer is a flexible framework that can infer cell types with various segmentation/detection models or manual annotations.  

\noindent\textbf{Computational efficiency of CGT.} Table \ref{tab_compute} displays the efficiency comparison between our proposed CGT and Hover-net. To evaluate the feasibility of real-world applications of CGT, we assessed the average parameter count (\#Para), inference time (Infer Time), and model size on ten whole slide images (WSIs). These WSIs have an average size of $36210 \times 71309$. `Ours' in Table~\ref{tab_compute} measures our method without including the segmentation tool. Our method increases 400+ MB storage and 25\% inference time when cooperating with existing segmentation methods, which is acceptable considering the low cost of the hard disk and the significant improvement of performance. 

\subsection{Ablation Study}
In Table \ref{table_abl},  we assess the strengths of CGT and the proposed Topology-Aware Pretraining (TAP) strategy. In the testing stage, all models adopt the binary segmentation masks generated by the same Hover-net. `UNet' denotes the feature extractor in CGT. `UNet ImageNet pretrained' is to initial the UNet with the ImageNet-1K pretrained weights. `Linear', `Transformer', `GCN' and `CGToken+Transformer' denote four classifiers: a linear embedding layer, a vanilla transformer without graph structure, a graph convolutional network and our proposed cell graph tokenization with a transformer encoder, respectively. 
For example, UNet Linear/Transformer/GCN pretrained means using Linear/Transformer/GCN as the classifier to pretrain the UNet for initialization. Our method is denoted as M8 where `UNet GCN pretrained' represents the TAP strategy.

\begin{figure}[!t]
\includegraphics[width=0.9\linewidth]{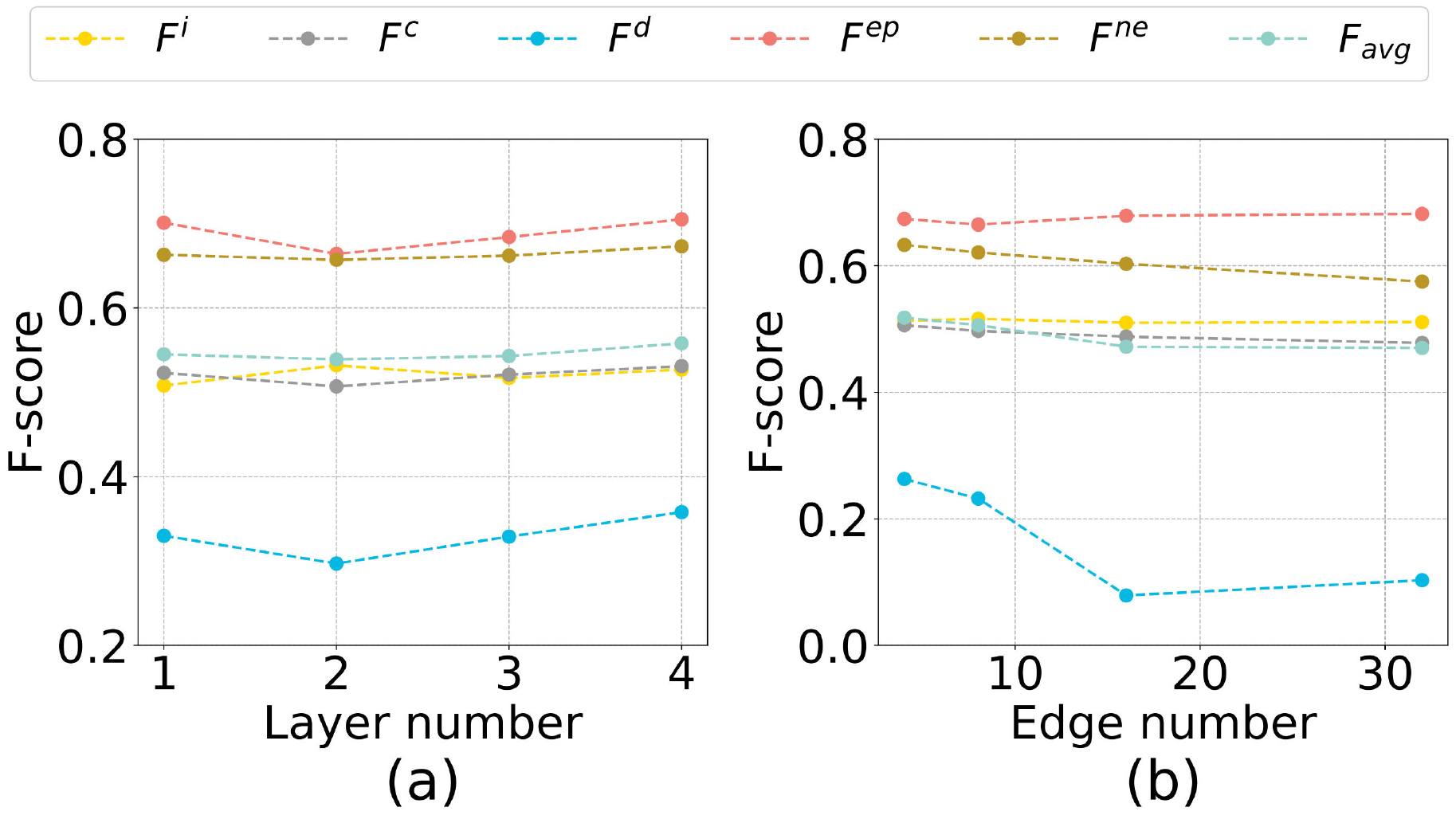}
\centering
\caption{Analysis of different choices for layer number ($L\in \{1,2,3,4\}$) in CGT and edge number ($E\in \{4,8,16,32\}$) in GCN (the topology-aware pretraining) on PanNuke dataset.}
\label{fig_hyper}
\end{figure}

\noindent \textbf{Effectiveness of the proposed pretraining.} 
To validate the proposed pretraining strategy, we compare M8 with M4-M6 and find that M8 using the TAP strategy significantly outperforms M4-M6 by 21\%, 13\% and 7.4\% in $F_{avg}$, respectively. The above results suggest that the TAP strategy using GCN is more effective than the simple ImageNet-pretraining, the pretraining guided by a linear layer and a vanilla transformer. We claim that it is because the TAP strategy makes the visual features aware of the graph connections and better adapt to our proposed CGT classifier. 

\noindent \textbf{Effectiveness of the CGT classifier.} 
Comparing M8 to M7 suggests that our proposed classifier built of cell graph tokenization and transformer encoder surpasses the vanilla transformer classifier without graph modeling by 4.6\% in $F_{avg}$. Comparing M5 to M1 and M8 to M3 shows that our proposed CGT classifier can outperform the linear and the GCN classifier by 7.9\% and 4\% in $F_{avg}$.
Besides, M1-M3 can be viewed as the combinations of the existing initialization and classifiers, and our overall method M8 exceeds these solutions by 4\%-21\% in $F_{avg}$. 

\noindent \textbf{Investigation of Hyper-parameters.}
In Figure \ref{fig_hyper}, we study the number of transformer layers $L$ of the CGT encoder on PanNuke dataset. If we set $L$ from 1 to 4, the average F-score first decrease by 0.6\% and then increases by 0.4\% and 1.5\%. The results indicate that the performance is improved slightly with increasing transformer layers. Due to the GPU memory limitation, we do not test for larger layer numbers.
The idea of our pretraining is that too dense connections with inferior features could result in unreasonable correlations and message passing, which affects the CGT training. In contrast, the GCN using a sparse graph is more robust. To verify the above idea, we use a denser graph for GCN-based pretraining by increasing edge number $E$. In Figure~\ref{fig_hyper}, as $E$ increases from 4 to 8 and 32, the average F-score of GCN does drop by 1.2\% and 4.8\%, which validates our assumption.

\section{Discussion}
\noindent\textbf{CGT vs. Transformer.}
The proposed CGT differs from vanilla transformers that compute correlations and pass messages between each pair of nodes equally. In contrast, the CGT defines edge features to describe pathological microenvironment and exploits link \& token markers to learn connections which emphasizes the attention between relevant cells. In Table~\ref{table_abl}, the CGT (M8) outperforms the vanilla transformer (M7/M2) by 4.6\%-11\% in $F_{avg}$, which indicates that the proposed CGT better models the cells and their interactions in pathological images than the vanilla transformers.

\noindent\textbf{GCN pretraining vs. Transformer pretraining.} We discuss why the pretraining with GCN as classifier (M8 in Table~\ref{table_abl}) is better than the one with vanilla transformer (M6). 
The GCN-pretraining adopts a sparse graph where the well-defined connections could guide the reasonable propagation of information. During the GCN-pretraining, the feature extractor is tuned by the gradients that are computed based on the well-defined edges and can adapt to the topology of cell graphs. However, in the transformer-pretraining, the gradients passed through the feature extractor are calculated from any pairs of nodes, even those that are irrelevant. Thus, the gradients in transformer-pretraining are more noisy and unreliable than those in GCN-pretraining, at the start of training. 

\section{Conclusion}
In this paper, a cell graph transformer (CGT) framework is proposed for identifying cell types with detected nucleus centroids. Our method embraces the transformer as a cell graph learner to fully exploit contexts and learn topological features. Both cell nodes and edges are viewed as input tokens to capture long-range correlations. The graph structure is embedded into the transformer encoder via link markers and token markers. Furthermore, we develop a novel topology-aware pretraining strategy that employs the robust local message-passing mechanism of graph convolutional networks to help pretrain the feature extractor of CGT. The experimental results display that the CGT model achieves the state-of-the-art nuclei classification performance on existing benchmarks. 

\section{Acknowledgments}
This work was supported in part by the National Natural Science Foundation of China (NO.~62102267, NO.~62322608), in part by the Guangdong Basic and Applied Basic Research Foundation (2023A1515011464), in part by the Shenzhen Science and Technology Program JCYJ20220818103001002), and in part by the Guangdong Provincial Key Laboratory of Big Data Computing, The Chinese University of Hong Kong, Shenzhen.

\begin{bibliography}{aaai24}
\end{bibliography}


\begin{thebibliography}{58}
\providecommand{\natexlab}[1]{#1}

\bibitem[{Abousamra et~al.(2021)Abousamra, Belinsky, Van~Arnam, Allard, Yee,
  Gupta, Kurc, Samaras, Saltz, and Chen}]{abousamra2021multi}
Abousamra, S.; Belinsky, D.; Van~Arnam, J.; Allard, F.; Yee, E.; Gupta, R.;
  Kurc, T.; Samaras, D.; Saltz, J.; and Chen, C. 2021.
\newblock Multi-class cell detection using spatial context representation.
\newblock In \emph{ICCV}, 4005--4014.

\bibitem[{Amgad et~al.(2022)Amgad, Atteya, Hussein, Mohammed, Hafiz, Elsebaie,
  Alhusseiny, AlMoslemany, Elmatboly, Pappalardo et~al.}]{amgad2022nucls}
Amgad, M.; Atteya, L.~A.; Hussein, H.; Mohammed, K.~H.; Hafiz, E.; Elsebaie,
  M.~A.; Alhusseiny, A.~M.; AlMoslemany, M.~A.; Elmatboly, A.~M.; Pappalardo,
  P.~A.; et~al. 2022.
\newblock NuCLS: A scalable crowdsourcing approach and dataset for nucleus
  classification and segmentation in breast cancer.
\newblock \emph{GigaScience}, 11.

\bibitem[{Anand, Gadiya, and Sethi(2020)}]{anand2020histographs}
Anand, D.; Gadiya, S.; and Sethi, A. 2020.
\newblock Histographs: graphs in histopathology.
\newblock In \emph{Medical Imaging 2020: Digital Pathology}, volume 11320,
  150--155. SPIE.

\bibitem[{Anklin et~al.(2021)Anklin, Pati, Jaume, Bozorgtabar,
  Foncubierta-Rodriguez, Thiran, Sibony, Gabrani, and
  Goksel}]{anklin2021learning}
Anklin, V.; Pati, P.; Jaume, G.; Bozorgtabar, B.; Foncubierta-Rodriguez, A.;
  Thiran, J.-P.; Sibony, M.; Gabrani, M.; and Goksel, O. 2021.
\newblock Learning whole-slide segmentation from inexact and incomplete labels
  using tissue graphs.
\newblock In \emph{MICCAI}, 636--646. Springer.

\bibitem[{Basha et~al.(2018)Basha, Ghosh, Babu, Dubey, Pulabaigari, and
  Mukherjee}]{basha2018rccnet}
Basha, S.~S.; Ghosh, S.; Babu, K.~K.; Dubey, S.~R.; Pulabaigari, V.; and
  Mukherjee, S. 2018.
\newblock Rccnet: An efficient convolutional neural network for histological
  routine colon cancer nuclei classification.
\newblock In \emph{ICARCV}, 1222--1227. IEEE.

\bibitem[{Chen et~al.(2021)Chen, Yang, Li, Zhao, Zha, and
  Wu}]{chen2021disentangle}
Chen, Z.; Yang, C.; Li, Q.; Zhao, F.; Zha, Z.-J.; and Wu, F. 2021.
\newblock Disentangle your dense object detector.
\newblock In \emph{ACM Multimedia}, 4939--4948.

\bibitem[{Cheng et~al.(2022)Cheng, Misra, Schwing, Kirillov, and
  Girdhar}]{cheng2022masked}
Cheng, B.; Misra, I.; Schwing, A.~G.; Kirillov, A.; and Girdhar, R. 2022.
\newblock Masked-attention mask transformer for universal image segmentation.
\newblock In \emph{CVPR}, 1290--1299.

\bibitem[{Demir, Gultekin, and Yener(2005)}]{demir2005augmented}
Demir, C.; Gultekin, S.~H.; and Yener, B. 2005.
\newblock Augmented cell-graphs for automated cancer diagnosis.
\newblock \emph{Bioinformatics}, 21(suppl\_2): ii7--ii12.

\bibitem[{Doan et~al.(2022)Doan, Song, Le~Vuong, Kim, and
  Kwak}]{doan2022sonnet}
Doan, T. N.~N.; Song, B.; Le~Vuong, T.~T.; Kim, K.; and Kwak, J.~T. 2022.
\newblock SONNET: A self-guided ordinal regression neural network for
  segmentation and classification of nuclei in large-scale multi-tissue
  histology images.
\newblock \emph{IEEE JBHI}.

\bibitem[{Dwivedi and Bresson(2020)}]{dwivedi2020generalization}
Dwivedi, V.~P.; and Bresson, X. 2020.
\newblock A generalization of transformer networks to graphs.
\newblock \emph{arXiv preprint arXiv:2012.09699}.

\bibitem[{Dwivedi et~al.(2020)Dwivedi, Joshi, Laurent, Bengio, and
  Bresson}]{dwivedi2020benchmarking}
Dwivedi, V.~P.; Joshi, C.~K.; Laurent, T.; Bengio, Y.; and Bresson, X. 2020.
\newblock Benchmarking graph neural networks.
\newblock \emph{arXiv preprint arXiv:2003.00982}.

\bibitem[{Fey and Lenssen(2019)}]{fey2019fast}
Fey, M.; and Lenssen, J.~E. 2019.
\newblock Fast graph representation learning with PyTorch Geometric.
\newblock \emph{ICLR Workshop}.

\bibitem[{Gamper et~al.(2020)Gamper, Koohbanani, Benes, Graham, Jahanifar,
  Khurram, Azam, Hewitt, and Rajpoot}]{gamper2020pannuke}
Gamper, J.; Koohbanani, N.~A.; Benes, K.; Graham, S.; Jahanifar, M.; Khurram,
  S.~A.; Azam, A.; Hewitt, K.; and Rajpoot, N. 2020.
\newblock Pannuke dataset extension, insights and baselines.
\newblock \emph{arXiv preprint arXiv:2003.10778}.

\bibitem[{Ge et~al.(2021)Ge, Liu, Wang, Li, and Sun}]{ge2021yolox}
Ge, Z.; Liu, S.; Wang, F.; Li, Z.; and Sun, J. 2021.
\newblock Yolox: Exceeding yolo series in 2021.
\newblock \emph{arXiv preprint arXiv:2107.08430}.

\bibitem[{Graham et~al.(2021)Graham, Jahanifar, Azam, Nimir, Tsang, Dodd, Hero,
  Sahota, Tank, Benes et~al.}]{graham2021lizard}
Graham, S.; Jahanifar, M.; Azam, A.; Nimir, M.; Tsang, Y.-W.; Dodd, K.; Hero,
  E.; Sahota, H.; Tank, A.; Benes, K.; et~al. 2021.
\newblock Lizard: A large-scale dataset for colonic nuclear instance
  segmentation and classification.
\newblock In \emph{ICCV Workshops}, 684--693.

\bibitem[{Graham et~al.(2019)Graham, Vu, Raza, Azam, Tsang, Kwak, and
  Rajpoot}]{graham2019hover}
Graham, S.; Vu, Q.~D.; Raza, S. E.~A.; Azam, A.; Tsang, Y.~W.; Kwak, J.~T.; and
  Rajpoot, N. 2019.
\newblock Hover-net: Simultaneous segmentation and classification of nuclei in
  multi-tissue histology images.
\newblock \emph{MIA}, 58: 101563.

\bibitem[{Grossman et~al.(2016)Grossman, Heath, Ferretti, Varmus, Lowy, Kibbe,
  and Staudt}]{grossman2016toward}
Grossman, R.~L.; Heath, A.~P.; Ferretti, V.; Varmus, H.~E.; Lowy, D.~R.; Kibbe,
  W.~A.; and Staudt, L.~M. 2016.
\newblock Toward a shared vision for cancer genomic data.
\newblock \emph{New England Journal of Medicine}, 375(12): 1109--1112.

\bibitem[{Guo et~al.(2023)Guo, Lu, Liu, Cheng, and Hu}]{guo2023visual}
Guo, M.-H.; Lu, C.-Z.; Liu, Z.-N.; Cheng, M.-M.; and Hu, S.-M. 2023.
\newblock Visual attention network.
\newblock \emph{Computational Visual Media}, 1--20.

\bibitem[{Hassan et~al.(2022)Hassan, Javed, Mahmood, Qaiser, Werghi, and
  Rajpoot}]{hassan2022nucleus}
Hassan, T.; Javed, S.; Mahmood, A.; Qaiser, T.; Werghi, N.; and Rajpoot, N.
  2022.
\newblock Nucleus Classification in Histology Images Using Message Passing
  Network.
\newblock \emph{MIA}, 102480.

\bibitem[{Huang et~al.(2023{\natexlab{a}})Huang, Li, Sun, Wan, and
  Li}]{huang2023prompt}
Huang, J.; Li, H.; Sun, W.; Wan, X.; and Li, G. 2023{\natexlab{a}}.
\newblock Prompt-based grouping transformer for nucleus detection and
  classification.
\newblock In \emph{MICCAI}, 569--579. Springer.

\bibitem[{Huang et~al.(2023{\natexlab{b}})Huang, Li, Wan, and
  Li}]{huang2023affine}
Huang, J.; Li, H.; Wan, X.; and Li, G. 2023{\natexlab{b}}.
\newblock Affine-Consistent Transformer for Multi-Class Cell Nuclei Detection.
\newblock In \emph{Proceedings of the IEEE/CVF International Conference on
  Computer Vision}, 21384--21393.

\bibitem[{Javed et~al.(2020)Javed, Mahmood, Fraz, Koohbanani, Benes, Tsang,
  Hewitt, Epstein, Snead, and Rajpoot}]{javed2020cellular}
Javed, S.; Mahmood, A.; Fraz, M.~M.; Koohbanani, N.~A.; Benes, K.; Tsang,
  Y.-W.; Hewitt, K.; Epstein, D.; Snead, D.; and Rajpoot, N. 2020.
\newblock Cellular community detection for tissue phenotyping in colorectal
  cancer histology images.
\newblock \emph{MIA}, 63: 101696.

\bibitem[{Kim et~al.(2022)Kim, Nguyen, Min, Cho, Lee, Lee, and Hong}]{kimpure}
Kim, J.; Nguyen, D.~T.; Min, S.; Cho, S.; Lee, M.; Lee, H.; and Hong, S. 2022.
\newblock Pure Transformers are Powerful Graph Learners.
\newblock \emph{NeurIPS}.

\bibitem[{Kim, Oh, and Hong(2021)}]{kim2021transformers}
Kim, J.; Oh, S.; and Hong, S. 2021.
\newblock Transformers generalize deepsets and can be extended to graphs \&
  hypergraphs.
\newblock \emph{NeurIPS}, 34: 28016--28028.

\bibitem[{Kirillov et~al.(2019)Kirillov, He, Girshick, Rother, and
  Doll{\'a}r}]{kirillov2019panoptic}
Kirillov, A.; He, K.; Girshick, R.; Rother, C.; and Doll{\'a}r, P. 2019.
\newblock Panoptic segmentation.
\newblock In \emph{CVPR}, 9404--9413.

\bibitem[{Kreuzer et~al.(2021)Kreuzer, Beaini, Hamilton, L{\'e}tourneau, and
  Tossou}]{kreuzer2021rethinking}
Kreuzer, D.; Beaini, D.; Hamilton, W.; L{\'e}tourneau, V.; and Tossou, P. 2021.
\newblock Rethinking graph transformers with spectral attention.
\newblock \emph{NeurIPS}, 34: 21618--21629.

\bibitem[{Krithiga and Geetha(2021)}]{krithiga2021breast}
Krithiga, R.; and Geetha, P. 2021.
\newblock Breast cancer detection, segmentation and classification on
  histopathology images analysis: a systematic review.
\newblock \emph{Archives of Computational Methods in Engineering}, 28:
  2607--2619.

\bibitem[{Lagree et~al.(2021)Lagree, Mohebpour, Meti, Saednia, Lu, Slodkowska,
  Gandhi, Rakovitch, Shenfield, Sadeghi-Naini et~al.}]{lagree2021review}
Lagree, A.; Mohebpour, M.; Meti, N.; Saednia, K.; Lu, F.-I.; Slodkowska, E.;
  Gandhi, S.; Rakovitch, E.; Shenfield, A.; Sadeghi-Naini, A.; et~al. 2021.
\newblock A review and comparison of breast tumor cell nuclei segmentation
  performances using deep convolutional neural networks.
\newblock \emph{Scientific Reports}, 11(1): 8025.

\bibitem[{Li et~al.(2021)Li, M{\"u}ller, Qian, Perez, Abualshour, Thabet, and
  Ghanem}]{li2021deepgcns}
Li, G.; M{\"u}ller, M.; Qian, G.; Perez, I. C.~D.; Abualshour, A.; Thabet,
  A.~K.; and Ghanem, B. 2021.
\newblock Deepgcns: Making gcns go as deep as cnns.
\newblock \emph{TPAMI}, 6923 -- 6939.

\bibitem[{Li et~al.(2020)Li, Xiong, Thabet, and Ghanem}]{li2020deepergcn}
Li, G.; Xiong, C.; Thabet, A.; and Ghanem, B. 2020.
\newblock Deepergcn: All you need to train deeper gcns.
\newblock \emph{arXiv preprint arXiv:2006.07739}.

\bibitem[{Li, Han, and Wu(2018)}]{li2018deeper}
Li, Q.; Han, Z.; and Wu, X.-M. 2018.
\newblock Deeper insights into graph convolutional networks for semi-supervised
  learning.
\newblock In \emph{AAAI}, volume~32.

\bibitem[{Lin, Wang, and Liu(2021)}]{lin2021mesh}
Lin, K.; Wang, L.; and Liu, Z. 2021.
\newblock Mesh graphormer.
\newblock In \emph{ICCV}, 12939--12948.

\bibitem[{Lin et~al.(2017)Lin, Doll{\'a}r, Girshick, He, Hariharan, and
  Belongie}]{lin2017feature}
Lin, T.-Y.; Doll{\'a}r, P.; Girshick, R.; He, K.; Hariharan, B.; and Belongie,
  S. 2017.
\newblock Feature pyramid networks for object detection.
\newblock In \emph{CVPR}, 2117--2125.

\bibitem[{Liu, Mundra, and Rajapakse(2011)}]{liu2011features}
Liu, S.; Mundra, P.~A.; and Rajapakse, J.~C. 2011.
\newblock Features for cells and nuclei classification.
\newblock In \emph{EMBC}, 6601--6604.

\bibitem[{Liu et~al.(2022{\natexlab{a}})Liu, Jia, Hou, Li, Zhang, Yan, Yang,
  Guo, Chen, Li et~al.}]{liu2022pathological}
Liu, Y.; Jia, Y.; Hou, C.; Li, N.; Zhang, N.; Yan, X.; Yang, L.; Guo, Y.; Chen,
  H.; Li, J.; et~al. 2022{\natexlab{a}}.
\newblock Pathological prognosis classification of patients with neuroblastoma
  using computational pathology analysis.
\newblock \emph{CBM}, 149: 105980.

\bibitem[{Liu et~al.(2021)Liu, Lin, Cao, Hu, Wei, Zhang, Lin, and
  Guo}]{liu2021swin}
Liu, Z.; Lin, Y.; Cao, Y.; Hu, H.; Wei, Y.; Zhang, Z.; Lin, S.; and Guo, B.
  2021.
\newblock Swin transformer: Hierarchical vision transformer using shifted
  windows.
\newblock In \emph{ICCV}, 10012--10022.

\bibitem[{Liu et~al.(2022{\natexlab{b}})Liu, Mao, Wu, Feichtenhofer, Darrell,
  and Xie}]{liu2022convnet}
Liu, Z.; Mao, H.; Wu, C.-Y.; Feichtenhofer, C.; Darrell, T.; and Xie, S.
  2022{\natexlab{b}}.
\newblock A convnet for the 2020s.
\newblock In \emph{CVPR}, 11976--11986.

\bibitem[{Lou et~al.(2022)Lou, Li, Li, Han, and Wan}]{lou2022pixel}
Lou, W.; Li, H.; Li, G.; Han, X.; and Wan, X. 2022.
\newblock Which pixel to annotate: a label-efficient nuclei segmentation
  framework.
\newblock \emph{IEEE Transactions on Medical Imaging}, 42(4): 947--958.

\bibitem[{Lou et~al.(2023{\natexlab{a}})Lou, Wan, Li, Lou, Li, Gao, and
  Li}]{lou2023structure}
Lou, W.; Wan, X.; Li, G.; Lou, X.; Li, C.; Gao, F.; and Li, H.
  2023{\natexlab{a}}.
\newblock Structure Embedded Nucleus Classification for Histopathology Images.
\newblock \emph{arXiv preprint arXiv:2302.11416}.

\bibitem[{Lou et~al.(2023{\natexlab{b}})Lou, Yu, Liu, Wan, Li, Liu, and
  Li}]{lou2023multi}
Lou, W.; Yu, X.; Liu, C.; Wan, X.; Li, G.; Liu, S.; and Li, H.
  2023{\natexlab{b}}.
\newblock Multi-stream Cell Segmentation with Low-level Cues for Multi-modality
  Images.
\newblock In \emph{Competitions in Neural Information Processing Systems},
  1--10. PMLR.

\bibitem[{Ma et~al.(2023)Ma, Xie, Ayyadhury, Ge, Gupta, Gupta, Gu, Zhang, Lee,
  Kim et~al.}]{ma2023multi}
Ma, J.; Xie, R.; Ayyadhury, S.; Ge, C.; Gupta, A.; Gupta, R.; Gu, S.; Zhang,
  Y.; Lee, G.; Kim, J.; et~al. 2023.
\newblock The Multi-modality Cell Segmentation Challenge: Towards Universal
  Solutions.
\newblock \emph{arXiv preprint arXiv:2308.05864}.

\bibitem[{Mahmood et~al.(2019)Mahmood, Borders, Chen, McKay, Salimian, Baras,
  and Durr}]{Mahmood2019Deep}
Mahmood, F.; Borders, D.; Chen, R.~J.; McKay, G.~N.; Salimian, K.~J.; Baras,
  A.; and Durr, N.~J. 2019.
\newblock Deep adversarial training for multi-organ nuclei segmentation in
  histopathology images.
\newblock \emph{IEEE TMI}, 39(11): 3257--3267.

\bibitem[{Oono and Suzuki(2020)}]{oono2020graph}
Oono, K.; and Suzuki, T. 2020.
\newblock Graph Neural Networks Exponentially Lose Expressive Power for Node
  Classification.
\newblock In \emph{ICLR}.

\bibitem[{Paszke et~al.(2017)Paszke, Gross, Chintala, Chanan, Yang, DeVito,
  Lin, Desmaison, Antiga, and Lerer}]{paszke2017automatic}
Paszke, A.; Gross, S.; Chintala, S.; Chanan, G.; Yang, E.; DeVito, Z.; Lin, Z.;
  Desmaison, A.; Antiga, L.; and Lerer, A. 2017.
\newblock Automatic differentiation in pytorch.
\newblock \emph{NIPS Workshop}.

\bibitem[{Pati et~al.(2022)Pati, Jaume, Foncubierta-Rodr{\'\i}guez, Feroce,
  Anniciello, Scognamiglio, Brancati, Fiche, Dubruc, Riccio
  et~al.}]{pati2022hierarchical}
Pati, P.; Jaume, G.; Foncubierta-Rodr{\'\i}guez, A.; Feroce, F.; Anniciello,
  A.~M.; Scognamiglio, G.; Brancati, N.; Fiche, M.; Dubruc, E.; Riccio, D.;
  et~al. 2022.
\newblock Hierarchical graph representations in digital pathology.
\newblock \emph{MIA}, 75: 102264.

\bibitem[{Schnorrenberg et~al.(1996)Schnorrenberg, Pattichis, Schizas,
  Kyriacou, and Vassiliou}]{schnorrenberg1996computer}
Schnorrenberg, F.; Pattichis, C.~S.; Schizas, C.~N.; Kyriacou, K.; and
  Vassiliou, M. 1996.
\newblock Computer-aided classification of breast cancer nuclei.
\newblock \emph{Technology and Health Care}, 4(2): 147--161.

\bibitem[{Sharma et~al.(2015)Sharma, Zerbe, Heim, Wienert, Behrens, Hellwich,
  and Hufnagl}]{sharma2015multi}
Sharma, H.; Zerbe, N.; Heim, D.; Wienert, S.; Behrens, H.-M.; Hellwich, O.; and
  Hufnagl, P. 2015.
\newblock A multi-resolution approach for combining visual information using
  nuclei segmentation and classification in histopathological images.
\newblock In \emph{VISAPP (3)}, 37--46.

\bibitem[{Sirinukunwattana et~al.(2017)Sirinukunwattana, Pluim, Chen, Qi, Heng,
  Guo, Wang, Matuszewski, Bruni, Sanchez et~al.}]{sirinukunwattana2017gland}
Sirinukunwattana, K.; Pluim, J.~P.; Chen, H.; Qi, X.; Heng, P.-A.; Guo, Y.~B.;
  Wang, L.~Y.; Matuszewski, B.~J.; Bruni, E.; Sanchez, U.; et~al. 2017.
\newblock Gland segmentation in colon histology images: The glas challenge
  contest.
\newblock \emph{MIA}, 35: 489--502.

\bibitem[{Vaswani et~al.(2017)Vaswani, Shazeer, Parmar, Uszkoreit, Jones,
  Gomez, Kaiser, and Polosukhin}]{vaswani2017attention}
Vaswani, A.; Shazeer, N.; Parmar, N.; Uszkoreit, J.; Jones, L.; Gomez, A.~N.;
  Kaiser, {\L}.; and Polosukhin, I. 2017.
\newblock Attention is all you need.
\newblock \emph{NIPS}, 30.

\bibitem[{Wei et~al.(2023)Wei, Xiang, Guanbin, Xiaoying, Chenghang, Feng, and
  Li}]{Wei2023Structure}
Wei, L.; Xiang, W.; Guanbin, L.; Xiaoying, L.; Chenghang, L.; Feng, G.; and Li,
  H. 2023.
\newblock Structure Embedded Nucleus Classification for Histopathology Images.
\newblock \emph{arXiv preprint arXiv:2302.11416}.

\bibitem[{Wu et~al.(2021)Wu, Jain, Wright, Mirhoseini, Gonzalez, and
  Stoica}]{wu2021representing}
Wu, Z.; Jain, P.; Wright, M.; Mirhoseini, A.; Gonzalez, J.~E.; and Stoica, I.
  2021.
\newblock Representing long-range context for graph neural networks with global
  attention.
\newblock \emph{NeurIPS}, 34: 13266--13279.

\bibitem[{Ying et~al.(2021)Ying, Cai, Luo, Zheng, Ke, He, Shen, and
  Liu}]{ying2021transformers}
Ying, C.; Cai, T.; Luo, S.; Zheng, S.; Ke, G.; He, D.; Shen, Y.; and Liu, T.-Y.
  2021.
\newblock Do transformers really perform badly for graph representation?
\newblock \emph{NeurIPS}, 34: 28877--28888.

\bibitem[{Yu et~al.(2023)Yu, Li, Lou, Liu, Wan, Chen, and Li}]{yu2023diffusion}
Yu, X.; Li, G.; Lou, W.; Liu, S.; Wan, X.; Chen, Y.; and Li, H. 2023.
\newblock Diffusion-based data augmentation for nuclei image segmentation.
\newblock In \emph{MICCAI}, 592--602. Springer.

\bibitem[{Zhang et~al.(2022)Zhang, Li, Liu, Zhang, Su, Zhu, Ni, and
  Shum}]{zhang2022dino}
Zhang, H.; Li, F.; Liu, S.; Zhang, L.; Su, H.; Zhu, J.; Ni, L.; and Shum, H.-Y.
  2022.
\newblock DINO: DETR with Improved DeNoising Anchor Boxes for End-to-End Object
  Detection.
\newblock In \emph{ICLR}.

\bibitem[{Zhang et~al.(2017)Zhang, Lu, Nogues, Summers, Liu, and
  Yao}]{zhang2017deeppap}
Zhang, L.; Lu, L.; Nogues, I.; Summers, R.~M.; Liu, S.; and Yao, J. 2017.
\newblock DeepPap: deep convolutional networks for cervical cell
  classification.
\newblock \emph{IEEE JBHI}, 21(6): 1633--1643.

\bibitem[{Zhao et~al.(2020)Zhao, Yang, Fang, Liu, Zhou, Zhang, Sun, Yang,
  Menze, Fan et~al.}]{zhao2020predicting}
Zhao, Y.; Yang, F.; Fang, Y.; Liu, H.; Zhou, N.; Zhang, J.; Sun, J.; Yang, S.;
  Menze, B.; Fan, X.; et~al. 2020.
\newblock Predicting lymph node metastasis using histopathological images based
  on multiple instance learning with deep graph convolution.
\newblock In \emph{CVPR}, 4837--4846.

\bibitem[{Zheng et~al.(2022)Zheng, Gindra, Green, Burks, Betke, Beane, and
  Kolachalama}]{zheng2022graph}
Zheng, Y.; Gindra, R.~H.; Green, E.~J.; Burks, E.~J.; Betke, M.; Beane, J.~E.;
  and Kolachalama, V.~B. 2022.
\newblock A graph-transformer for whole slide image classification.
\newblock \emph{IEEE TMI}, 41(11): 3003--3015.

\bibitem[{Zhou et~al.(2019)Zhou, Graham, Alemi~Koohbanani, Shaban, Heng, and
  Rajpoot}]{zhou2019cgc}
Zhou, Y.; Graham, S.; Alemi~Koohbanani, N.; Shaban, M.; Heng, P.-A.; and
  Rajpoot, N. 2019.
\newblock Cgc-net: Cell graph convolutional network for grading of colorectal
  cancer histology images.
\newblock In \emph{ICCV Workshops}.

\end{thebibliography}
\end{document}